\documentclass{article}
\usepackage{arxiv}

\usepackage[english]{babel}
\usepackage[utf8]{inputenc}  

\usepackage{amsmath, amssymb, amsfonts}
\usepackage{amsthm}
\usepackage{mathtools}
\usepackage{cases}
\usepackage{bm}              
\usepackage{empheq}
\usepackage{systeme}
\usepackage{makecell}

\expandafter\let\csname equation*\endcsname\relax
\expandafter\let\csname endequation*\endcsname\relax

\usepackage{algorithm}
\usepackage{algpseudocode}

\usepackage{graphicx}
\usepackage{adjustbox}
\usepackage{subcaption}
\usepackage{float}
\usepackage{mwe}             

\usepackage{booktabs}
\usepackage{arydshln}
\usepackage{multirow}
\usepackage{array}

\usepackage{textcomp}
\usepackage{xcolor}
\usepackage{soul}
\usepackage{tipa}            

\usepackage{tikz}
\usepackage{xparse}
\usetikzlibrary{decorations.pathreplacing}
\usetikzlibrary{spy,calc}

\usepackage{listings}

\usepackage[
    bookmarksopen,
    bookmarksdepth=2,
    breaklinks=true
]{hyperref}
\usepackage{cleveref}

\usepackage[title]{appendix}
\usepackage{manyfoot}

\usepackage{comment}
\usepackage{xparse}
\usepackage{url}

\newcommand{\RR}{\mathbb{R}}
\newcommand{\G}{\mathcal{G}}

\newcommand{\vK}{\boldsymbol{K}}
\newcommand{\vI}{\boldsymbol{I}}
\newcommand{\ve}{\boldsymbol{e}}
\newcommand{\vx}{\boldsymbol{x}}
\newcommand{\vy}{\boldsymbol{y}}
\newcommand{\vz}{\boldsymbol{z}}
\newcommand{\vepsilon}{\boldsymbol{\epsilon}}

\makeatletter
\def\adl@drawiv#1#2#3{%
        \hskip.5\tabcolsep
        \xleaders#3{#2.5\@tempdimb #1{1}#2.5\@tempdimb}%
                #2\z@ plus1fil minus1fil\relax
        \hskip.5\tabcolsep}
\newcommand{\cdashlinelr}[1]{%
  \noalign{\vskip\aboverulesep
           \global\let\@dashdrawstore\adl@draw
           \global\let\adl@draw\adl@drawiv}
  \cdashline{#1}
  \noalign{\global\let\adl@draw\@dashdrawstore
           \vskip\belowrulesep}}
\makeatother

\usepackage{tikz}
\usepackage{calc} 
\usetikzlibrary{calc} 
\usetikzlibrary{positioning}

\definecolor{ZoomRed}{RGB}{255,0,0}

\newcommand{\imgwithzoom}[7]{%
\begin{tikzpicture}

    \node[anchor=south west, inner sep=0] (MainImg) at (0,0) {\includegraphics[#1]{#2}};
    
    \path (MainImg.south west); \pgfgetlastxy{\Mx}{\My}
    \path (MainImg.north east); \pgfgetlastxy{\Mw}{\Mh}
    
    \pgfmathsetmacro{\SourceBoxFrac}{#5} 
    \pgfmathsetmacro{\InsetBoxFrac}{#6}  
    
    \pgfmathsetmacro{\SourceSize}{\SourceBoxFrac * \Mw}
    \pgfmathsetmacro{\InsetSize}{\InsetBoxFrac * \Mw}
    
    \pgfmathsetmacro{\Mag}{\InsetSize / \SourceSize}
    
    \pgfmathsetmacro{\CenterRelX}{#3}
    \pgfmathsetmacro{\CenterRelY}{#4}
    
    \pgfmathsetmacro{\Cx}{\CenterRelX * \Mw}
    \pgfmathsetmacro{\Cy}{\CenterRelY * \Mh}
    
    \coordinate (SrcCenter) at (\Cx pt, \Cy pt);
    \coordinate (SrcSW) at (\Cx pt - 0.5*\SourceSize pt, \Cy pt - 0.5*\SourceSize pt);
    \coordinate (SrcNE) at (\Cx pt + 0.5*\SourceSize pt, \Cy pt + 0.5*\SourceSize pt);
    
    \def\BorderPad{0.01*\Mw} 
    \coordinate (InsetSW) at (\Mw - \BorderPad - \InsetSize pt, \BorderPad);
    \coordinate (InsetNE) at (\Mw - \BorderPad, \BorderPad + \InsetSize pt);
    \coordinate (InsetCenter) at ($ (InsetSW)!0.5!(InsetNE) $);

    \begin{scope}
        \clip (InsetSW) rectangle (InsetNE);
        
        \fill[white] (InsetSW) rectangle (InsetNE);
        
        \pgfmathsetmacro{\ShiftX}{\Cx * \Mag}
        \pgfmathsetmacro{\ShiftY}{\Cy * \Mag}
        
        \coordinate (PlacementPoint) at ($ (InsetCenter) + (-\ShiftX pt, -\ShiftY pt) $);
        
        \node[anchor=south west, inner sep=0, scale=\Mag] at (PlacementPoint) 
             {\includegraphics[#1]{#2}};
    \end{scope}

    \ifnum#7=1
        \draw[ZoomRed, line width=1pt] (SrcSW) rectangle (SrcNE);
    \fi
    
    \draw[ZoomRed, line width=1pt] (InsetSW) rectangle (InsetNE);

\end{tikzpicture}%
}

\newcommand{\imgmetric}[4]{
  \begin{tikzpicture}[baseline=(img.base)]
    \node[inner sep=0] (img) {\includegraphics[width=#4]{#1}};
    \node[
      anchor=north west,
      xshift=1.5pt, yshift=-1.5pt,
      fill=white, fill opacity=0.75,
      text opacity=1,
      inner sep=1.2pt,
      rounded corners=1pt,
      font=\tiny
    ] at (img.north west) {\shortstack[l]{PSNR: #2\\SSIM: #3}};
  \end{tikzpicture}%
}

\title{A Diffusion-Based Generative Prior Approach to Sparse-view Computed Tomography}

\author{
 Davide Evangelista \\
  Department of Computer Science and Engineering\\
  University of Bologna\\
  Mura Anteo Zamboni, 7, 40126, Italy \\
  \texttt{davide.evangelista5@unibo.it} \\
   \And
 Pasquale Cascarano \\
  Department of the Art\\
  University of Bologna\\
  via Barberia, 4, 40123, Italy \\
  \texttt{pasquale.cascarano2@unibo.it}
  \And
 Elena Loli Piccolomini \\
  Department of Computer Science and Engineering\\
  University of Bologna\\
  Mura Anteo Zamboni, 7, 40126, Italy \\
  \texttt{elena.loli@unibo.it}
}
\date{}

\begin{document}

\maketitle

\begin{abstract}
The reconstruction of X-rays CT images from sparse or limited-angle geometries is a highly challenging task. The lack of data typically results in artifacts in the reconstructed image and may even lead to object distortions.
For this reason, the use of deep generative models in this context has great interest and potential success.
In the Deep Generative Prior (DGP) framework, the use of diffusion-based generative models is combined with an iterative optimization algorithm for the reconstruction of CT images from sinograms acquired under sparse geometries, to maintain the explainability of a model-based approach while introducing the generative power of a neural network.
There are therefore several aspects that can be further investigated within these frameworks to improve reconstruction quality, such as image generation, the model, and the iterative algorithm used to solve the minimization problem, for which we propose modifications with respect to existing approaches. The results obtained even under highly sparse geometries are very promising, although further research is clearly needed in this direction.
\end{abstract}


\section{Introduction}\label{sec1}
The problem of reconstructing X-ray CT images from very sparse-view data has gained increasing attention in medical imaging research, driven by the need to reduce the radiation dose delivered to patients while preserving reconstruction accuracy~\cite{withers2021x}. 
Like many inverse problems, sparse-view CT reconstruction is ill-posed, meaning that solutions may be non-unique, unstable, or highly sensitive to noise~\cite{bertero2021introduction}.
A widely adopted strategy to address this issue consists of computing an approximate solution via a variational regularization approach, where the reconstructed image is obtained by solving an optimization problem that combines a data-fidelity term (typically based on a least-squares formulation) with a regularization term that enforces \emph{a priori} structural assumptions on the solution.
The main advantage of this framework lies in its explicit and interpretable consistency between the reconstructed image and the measured data, supported by solid mathematical foundations. However, it also suffers from notable drawbacks, including high computational cost and the need for careful tuning of multiple hyperparameters~\cite{liu2013total,kim2014sparse,10415495}.

Over the past decade, deep learning approaches based on Convolutional Neural Networks (CNNs) have demonstrated superior performance compared to classical variational methods in solving linear inverse problems, including CT reconstruction~\cite{arridge2019solving,wang2020deep,mccann2017convolutional,cascarano2022plug,evangelista2023rising}. Nevertheless, CNN-based methods often exhibit a significant degradation in performance in highly noisy \cite{evangelista2025or} or highly sparse-view \cite{zhang2026msdiff} regimes, due to their reliance on local spatial correlations that are severely compromised under extreme data undersampling.
In addition, in medical imaging, it is often difficult to obtain large paired datasets consisting of input measurements and corresponding ground-truth images, which limits the applicability of supervised learning approaches \cite{morotti2021green}.

For these reasons, recent research has increasingly focused on generative and unsupervised models, which do not require paired training data, and in particular on diffusion-based generative models~\cite{kazerouni2023diffusion,webber2024diffusion,wang2025diffusion}. 
While highly powerful diffusion models pre-trained on large-scale photographic image datasets are publicly available, this is generally not the case in the medical imaging domain. As a result, medical image generation typically relies on substantially smaller datasets, leading to models that do not yet match the quality achieved in natural image synthesis. This limitation remains a key challenge for the reliable deployment of generative models in clinical applications. 
Despite this, research activity in the use of generative models for solving linear inverse problems has grown rapidly~\cite{pan2020physics,duff2024regularising}.
Several works have proposed to enforce data consistency within diffusion models by solving an optimization problem at each step of the reverse diffusion process, thereby ensuring agreement between generated images and observed measurements~\cite{song2021solving,chung2022improving,chung2023solving}. 
A specific line of research is represented by the Deep Generative Prior (DGP) framework, originally introduced in~\cite{bora2017compressed} and later extended in~\cite{duff2024regularising,pan2021exploiting} using Variational Autoencoders (VAEs)~\cite{asperti2021survey} and Generative Adversarial Networks (GANs)~\cite{goodfellow2014generative} for various imaging inverse problems.
The core idea behind DGPs is to use a deep generative model as an implicit prior to regularize an inverse problem, while preserving a model-based and interpretable formulation.
Within the DGP framework, the solution of the inverse problem is generated directly by a pre-trained generator, while data consistency is enforced through an optimization problem that constrains the generated image to match the observed measurements. In essence, instead of imposing an explicit prior (such as total variation or sparsity), the solution is constrained to lie in the range of a generative model that has learned the distribution of plausible images. This is especially important in medical imaging, where a clear decoupling of physical modeling and prior information improves interpretability relative to end-to-end CNN methods.
However, previous studies~\cite{bora2017compressed,duff2024regularising,pan2021exploiting} have reported suboptimal reconstruction quality, as the performance of DGPs critically depends both on the expressive power of the generator and on the ability of the optimization procedure to escape poor local minima in a highly non-convex loss landscape.\\
In this work, we consider a Deep Generative Prior (DGP) formulation for sparse-view CT reconstruction in which a Denoising Diffusion Implicit Model (DDIM)~\cite{DDIM} is employed as the generative model. 
The inverse problem is formulated as the minimization of a least-squares data-fidelity term, regularized by a Tikhonov penalty on the latent variable, as suggested by the maximum a posteriori (MAP) interpretation, and a Total Variation (TV) regularizer acting on the generated image, in line with classical approaches in medical image reconstruction. While this formulation itself is not novel \cite{wang2024dmplug}, our focus is on addressing the practical optimization challenges that arise when applying DGPs to severely ill-posed CT reconstruction problems.

The main contributions of this paper are as follows: \textit{(i)} we propose a physics-informed initialization strategy that significantly improves optimization in the DGP framework. Specifically, we compute a coarse reconstruction using Filtered Backprojection (FBP) from the sparse-view sinogram, and then perform DDIM inversion to obtain the corresponding latent representation. This latent vector is used to initialize the DGP optimization, enabling the algorithm to escape poor local minima and consistently recover higher-quality reconstructions; \textit{(ii)} we introduce a dynamic learning-rate schedule for DGP optimization based on cosine annealing. We empirically demonstrate that progressively reducing the step size during the optimization leads to improved convergence behavior and superior reconstruction quality compared to fixed learning-rate schemes; 
and \textit{(iii)} we publicly release a diffusion model pre-trained on chest CT images using an augmented version of the widely-known Mayo's dataset. This model is intended to support open and reproducible research, and to facilitate future work on diffusion-based priors for CT reconstruction. The model weights, together with the code to replicate the experiments discussed in this paper, are available at: \url{https://github.com/devangelista2/RD-DGP}. 

The paper is organized as follows. Section \ref{sec:related_works} reviews related work and highlights the key differences of our proposed approach. Section \ref{sec:method} presents a detailed description of the proposed model and the iterative scheme adopted.
Section \ref{sec:background} introduces the experimental setup. 
Section \ref{sec:numexp} discusses the numerical results, and finally, Section \ref{sec:concl} offers concluding remarks.

\section{Related Works \label{sec:related_works}}

The use of the DGP framework and DDIMs has already been investigated in previous works, although a gap remains in the literature regarding approaches that effectively combine both.

In~\cite{dhar2018modeling}, the authors employ Variational Autoencoders (VAEs) as generative models alongside a single $\ell_1$ regularization function, with a focus on image deblurring. In~\cite{duff2024regularising}, a GAN-based generative model is used with $\ell_2$ regularization, and the numerical tests are conducted on small geometric images for CT reconstruction. In both studies, the suboptimal performance of the VAEs and GANs negatively impacted the final outcomes.
In contrast to these approaches, our work employs a generator based on diffusion models~\cite{DDIM,asperti2023image}. Diffusion models have emerged as a powerful paradigm, due to their ability to capture complex data distributions, providing optimal performances while avoiding the common drawbacks of other generative models, such as mode collapse~\cite{kossale2022mode,xia2023diffir}, as shown in \cite{rombach2022high}.

One of the current main research directions focuses on how to guide the generation process toward solutions that are consistent with the given measurements.
Some works achieve this by projecting intermediate solutions onto the feasible set or enforcing hard constraints~\cite{song2023solving, chung2022improving}. For instance,~\cite{song2023solving} performs such projections by solving an optimization problem during reverse sampling to enforce measurement consistency.
Other approaches guide the generation iteratively through gradient-based updates, enabling a gradual alignment with the measurements~\cite{luo2023refusion, xia2023diffir, chung2022diffusion}. For example,~\cite{chung2022diffusion} guides reverse diffusion by adjusting the score with a likelihood-informed gradient computed via Tweedie’s estimate of the clean image, achieving consistency without explicit projections and avoiding noise amplification.
Some recent works also combine these strategies to strike a balance between reconstruction accuracy and computational efficiency~\cite{garber2024image, zhu2023denoising}.
None of the aforementioned approaches explicitly adopt the DGP framework as we propose in this work. Our formulation allows for greater flexibility in enforcing both data consistency and incorporating different types of regularization. 

In the recent study~\cite{wang2024dmplug}, a  model, named DMPlug, combining the DGP framework with Diffusion Models is proposed for image restoration. 
Our work is mainly inspired by this article, in which it is shown that the fit using the available data is more accurate.
It differs from the previous approach~\cite{wang2024dmplug}  in several aspects: we propose a different initialization of the inverse diffusion process, an adaptive learning rate strategy, the inclusion of two regularization terms in the objective function, and we apply the method to a tomographic reconstruction problem, where the data are known to reside in a space different from that of the image. we will compare our method  with DMPlug in the section on numerical results.

\section{Methods\label{sec:method}}

\begin{figure}[ht]
    \centering
    \includegraphics[width=0.8\linewidth]{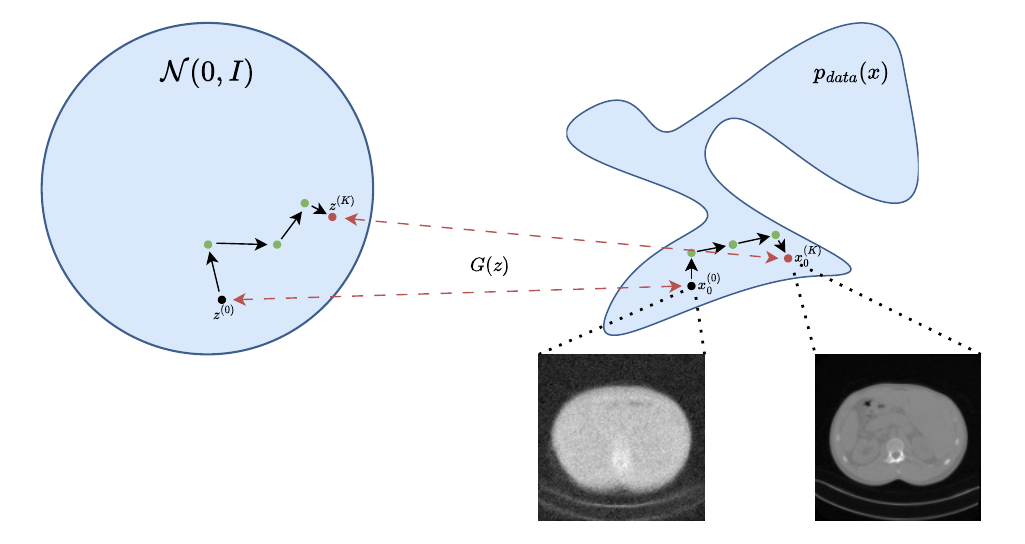}
    \caption{Graphical illustration of the Deep Generative Prior approach with Denoising Diffusion Implicit Models. It illustrates how to generate a single iterate $x_0^{K}$ of the optimization method  after $K$ reverse diffusion steps starting from $x_0$.}
    \label{fig:graphical_abstract}
\end{figure}

In this section, we present our regularized Deep Generative Prior framework for CT image reconstruction based on diffusion models.
We begin by introducing the inverse problem setting and reviewing DGPs; then, we describe diffusion-based generative models,  which we employ to parametrize the space of admissible reconstructions.
These components lead to a regularized latent optimization problem, which forms the basis of the proposed Regularized Diffusion-based Deep Generative Prior (RD-DGP) algorithm. A graphical overview of the overall approach is reported in Fig.~\ref{fig:graphical_abstract}. 

We consider CT image reconstruction as a linear inverse problem, where the goal is to recover an image $\vx \in \RR^n$ from noisy measurements $\vy^\delta \in \RR^m$ according to
\begin{equation}\label{inverse-problem}
 \vy^\delta = \vK \vx + \ve,
\end{equation}
where $\vK \in \RR^{m \times n}$ denotes the known forward operator describing the projection process, and $\ve \in \RR^m$ represents additive noise, with $\delta>0$ indicating the noise level.

\subsection{Deep Generative Prior\label{sub:model}}

Given a generator $\G(\vz)$ with latent variable $\vz \in \mathcal{Z}$, where $\mathcal{Z}$ denotes the latent space, image reconstruction can be formulated as the problem of finding a latent code whose generated image is consistent with the observed measurements, leading to
\begin{equation}\label{DGP-formulation}
    \vz^* \in \arg\min_{\vz \in \mathcal{Z}} \frac{1}{2}||\vK \G(\vz) - \vy^\delta||_2^2,
    \quad
    \vx^* = \G(\vz^*).
\end{equation}

When $\G$ is represented by a deep neural network, the objective in~\eqref{DGP-formulation} is highly non-convex and the reconstruction problem becomes ill-posed.
As a consequence, solutions obtained from the unregularized formulation may be sensitive to noise, initialization, and local minima. \\
To improve robustness and incorporate additional prior information, we consider a regularized latent optimization framework, in which constraints are imposed both in the latent space and in the image domain.
This leads to the following regularized formulation:
\begin{align}\label{eq:RD-DGP}
\vz^*
\in \arg\min_{\vz \in \mathcal{Z}} \mathcal{F}(\vz;\vy^\delta)
:=
\frac{1}{2}||\vK \G(\vz) - \vy^\delta||_2^2
+ \lambda_1 \mathcal{R}_1(\vz)
+ \lambda_2 \mathcal{R}_2(\G(\vz)).
\end{align}
Here, the data-fidelity term enforces consistency with the measurements, while the regularizers $\mathcal{R}_1$ and $\mathcal{R}_2$ promote adherence to the latent prior distribution and impose desired structural properties on the reconstructed image, respectively.

Since deep generative models are typically trained assuming a standard Gaussian prior on the latent variable $\vz \sim \mathcal{N}(\boldsymbol{0},\vI)$, a Maximum-A-Posteriori (MAP) interpretation naturally suggests choosing
\begin{equation}\label{eq:reg_tik}
\mathcal{R}_1(\vz) := -\log p(\vz) = ||\vz||_2^2.
\end{equation}
The image-domain regularizer $\mathcal{R}_2(\G(\vz))$ is introduced to enforce prior structural properties such as piecewise smoothness and sharp anatomical boundaries.
In this work, we adopt a smoothed, continuously differentiable version of Total Variation regularization, which is well suited to CT imaging due to its ability to promote piecewise-constant structures.

\subsection{Denoising Diffusion Implicit Models}

In this work, the generator $\G$ is instantiated using a diffusion-based generative model.
Specifically, we consider a Denoising Diffusion Probabilistic Model (DDPM) trained for image synthesis and we employ its deterministic Denoising Diffusion Implicit Model (DDIM) formulation for sampling.

Diffusion models learn the data distribution by progressively corrupting training samples with Gaussian noise and learning to reverse this process.
Starting from a clean image $\vx_0 \sim p_{\text{data}}$, the forward diffusion process is defined as
\begin{equation}\label{eq:forward_diffusion}
\vx_t = \sqrt{\alpha_t}\,\vx_0 + \sqrt{1-\alpha_t}\,\boldsymbol{\epsilon}_t,
\qquad
\boldsymbol{\epsilon}_t \sim \mathcal{N}(\boldsymbol{0},\vI),
\end{equation}
where $\{\alpha_t\}_{t=0}^T \subseteq [0, 1]$ is a monotonic decreasing diffusion schedule satisfying $\alpha_T \approx 0$.
Consequently, as $t$ increases, $\vx_t$ converges in distribution to standard Gaussian noise.

The reverse diffusion process is learned by training a neural network $f_\Theta(\vx_t,t)$ to predict the injected noise at each diffusion step.
DDIMs~\cite{DDIM} define a deterministic, non-Markovian reverse process that maps an initial noise realization $\vz \sim \mathcal{N}(\boldsymbol{0},\vI)$ to an image sample without introducing additional randomness.
As a result, the reverse diffusion process can be interpreted as a deterministic generator
\begin{equation}
\vx_0 = \G(\vz),
\end{equation}
which maps a latent code to an image lying on the learned data manifold.

This deterministic parametrization is particularly suitable for inverse problems, as it enables gradient-based optimization directly in the latent space.
The regularized formulation in~\eqref{eq:RD-DGP}, together with the DDIM-based generator $\G$, defines the foundation of the reconstruction approach proposed in this work.

\subsection{The RD-DGP algorithm \label{sub:alg}}

In this section, we describe the proposed RD-DGP algorithm and its main implementation details.
RD-DGP solves the regularized latent optimization problem in~\eqref{eq:RD-DGP}, where the diffusion-based generator $\G$ parametrizes the space of admissible reconstructions through the latent variable $\vz$.
The optimization is performed in the latent space, while both the data-fidelity term and the image-domain regularizer are evaluated after mapping to the image domain via $\vx_0=\G(\vz)$.
A graphical summary of the overall pipeline is reported in Fig.~\ref{fig:graphical_abstract}, while the corresponding pseudocode is provided in Algorithm~\ref{alg:RD-DGP}.

The minimization problem in~\eqref{eq:RD-DGP} is solved using Adam. At each iteration $k$, the latent variable is updated by descending the gradient of the objective functional $\mathcal{F}(\vz;\vy^\delta)$ according to
\begin{align}
\vz^{(k+1)} = \vz^{(k)} - \nu_k \nabla_{\vz}\mathcal{F}(\vz^{(k)};\vy^\delta),
\end{align}
where $\nu_k>0$ is the step size selected by Adam, and $\nabla_{\vz}\mathcal{F}(\vz^{(k)};\vy^\delta)$ is computed via backpropagation through the generator $\G$.
Given the current latent iterate $\vz^{(k)}$, the corresponding reconstruction is obtained as $\vx_0^{(k)}=\G(\vz^{(k)})$. 

Due to the non-convexity of $\mathcal{F}(\vz;\vy^\delta)$, induced by the deep network defining $\G$, and the ill-posedness of the operator $\vK$, initialization has a strong impact on both convergence and reconstruction quality.
Purely random initializations may drive the optimization toward poor local minima, leading to reconstructions that are either insufficiently data-consistent or weakly aligned with the learned diffusion prior.
To mitigate this issue, we adopt a problem-adaptive initialization strategy based on deterministic DDIM inversion.

Given the measured sinogram $\vy^\delta$, we first compute a filtered back-projection reconstruction~\cite{kak2001principles}
\begin{align}
\vx_0^{\mathrm{FBP}} = \mathrm{FBP}(\vy^\delta),
\end{align}
which provides a coarse but data-consistent estimate of the target image.
We then map $\vx_0^{\mathrm{FBP}}$ into the diffusion latent space by applying deterministic DDIM inversion under the \emph{same} diffusion schedule $\{\alpha_t\}_{t=0}^T$ used during training.
Starting from $\vx_0^{\mathrm{FBP}}$, the inversion constructs a sequence $\{\vx_t^{\mathrm{FBP}}\}_{t=0}^T$ via
\begin{align*}
\hat{\boldsymbol{\epsilon}}_t^{\mathrm{FBP}}  &= f_\Theta(\vx_{t-1}^{\mathrm{FBP}},t-1), \\
\hat{\vx}_0^{\mathrm{FBP}}  &= \frac{\vx_{t-1}^{\mathrm{FBP}} - \sqrt{1-\alpha_{t-1}}\,\hat{\boldsymbol{\epsilon}}_t^{\mathrm{FBP}} }{\sqrt{\alpha_{t-1}}}, \\
\vx_t^{\mathrm{FBP}}  &= \sqrt{\alpha_t}\,\hat{\vx}_0^{\mathrm{FBP}} + \sqrt{1-\alpha_t}\,\hat{\boldsymbol{\epsilon}}_t^{\mathrm{FBP}},
\end{align*}
for $t=1,\dots,T$.
At the final step, we obtain a latent representation $\vz_{\mathrm{FBP}} := \vx_T^{\mathrm{FBP}}$ such that $\G(\vz^{\mathrm{FBP}})\approx \vx_0^{\mathrm{FBP}}$.
We then initialize the latent optimization with $\vz^{(0)}=\vz^{\mathrm{FBP}}$.

To improve robustness and convergence, we additionally employ a cosine annealing schedule for the Adam step size.
For $k=0,\dots,\texttt{maxit}$, the learning rate is set as
\begin{align}\label{eq:cosine_schedule}
\nu_k
=
\nu_{\min}
+
\frac{1}{2}\left(\nu_{\max}-\nu_{\min}\right)
\left(
1 + \cos \left(\frac{\pi k}{\texttt{maxit}}\right)
\right),
\end{align}
with $\nu_{\max}=10^{-2}$ and $\nu_{\min}=10^{-5}$.
This schedule enables larger updates in the early iterations, supporting exploration of the latent space, while progressively reducing the step size to allow fine-scale refinement.

Overall, the combination of DDIM-inversion initialization and cosine-annealed latent optimization improves the stability of RD-DGP and reduces the tendency to converge to suboptimal local minima.
As shown in the experimental section, this strategy consistently yields reconstructions that are both data-consistent and well aligned with the learned diffusion prior.

\section{Experimental setup \label{sec:background}}
In this section, we describe the experimental pipeline used to validate our approach, including the data preparation process, the architecture and training details of the diffusion model, and the reconstruction setting adopted for evaluation.

\subsection{Data preparation}
The dataset used for training is the Mayo Clinic dataset~\cite{mccollough2016tu}, consisting of 3305 images with $512 \times 512$ pixels of human abdomen, that we have down-scaled to $256 \times 256$ pixels and normalized in the range $[0, 1]$.
For each ground truth image $\vx^{GT}$ in the dataset, we have simulated noisy projections $\vy^{\delta}$ as:
\begin{align}
    \vy^\delta = \vK \vx^{GT} + \delta || \vK \vx^{GT} ||_\infty \ve, \qquad \ve \sim \mathcal{N}(\boldsymbol{0}, I)
\end{align}
The matrix $\vK$ has been implemented with ASTRA Toolbox~\cite{van2015astra} functions, simulating a sparse parallel beam geometry with $n_{\alpha}$ projections uniformly distributed within the angular range $\Gamma=[0^\circ,180^\circ]$, with a detector resolution of $2 \sqrt{256}$ pixels.

\begin{algorithm}[H]
\centering
\caption{RD-DGP scheme}\label{alg:RD-DGP}
\begin{algorithmic}[1]
\Require observations $\vy^\delta$, pre-trained DDIM model $f_\Theta(\vx_t,t)$, diffusion schedule $\{\alpha_t\}_{t=0}^{T}$

\Statex \texttt{// Initialization}
\State \textbf{compute} $\vx^{\mathrm{FBP}}_0 := \mathrm{FBP}(\vy^\delta)$
\For{$t = 1,\dots,T$}
    \State $\hat{\boldsymbol{\epsilon}}_t^{FBP} = f_\Theta(\vx_{t-1}^{FBP},t-1)$
    \State $\hat{\vx}_0^{FBP} =
    \dfrac{\vx_{t-1}^{FBP} - \sqrt{1-\alpha_{t-1}}\,\hat{\boldsymbol{\epsilon}}_t^{FBP}}{\sqrt{\alpha_{t-1}}}$
    \State $\vx_{t}^{FBP} =
    \sqrt{\alpha_t}\,\hat{\vx}_0^{FBP} +
    \sqrt{1-\alpha_t}\,\hat{\boldsymbol{\epsilon}}_t^{FBP}$
\EndFor

\Statex
\Statex \texttt{// Reconstruction}
\State $\vz^{(0)} := \vx_T^{FBP}$, $k=0$, $\nu_0 > 0$

\For{$k = 1,\dots,\texttt{maxit}$}
    \State \textbf{set} $\vx_T^{(k)} = \vz^{(k)}$
    \For{$t = T,\dots,1$}
        \State $\hat{\boldsymbol{\epsilon}}_t^{(k)} = f_\Theta(\vx_t^{(k)},t)$
        \State $\hat{\vx}_0^{(k)} =
        \dfrac{\vx_t^{(k)} - \sqrt{1-\alpha_t}\,\hat{\boldsymbol{\epsilon}}_t^{(k)}}{\sqrt{\alpha_t}}$
        \State $\vx_{t-1}^{(k)} =
        \sqrt{\alpha_{t-1}}\,\hat{\vx}_0^{(k)} +
        \sqrt{1-\alpha_{t-1}}\,\hat{\boldsymbol{\epsilon}}_t^{(k)}$
    \EndFor
    \State \textbf{assign} $\vx^{(k)} \gets \vx_0^{(k)}$

    \Statex
    \State \textbf{compute} $\mathcal{F}_k := \mathcal{F}(\vz^{(k)},\vy^\delta)$ according to \eqref{eq:RD-DGP}
    \State \textbf{update} $\nu_k$ according to \eqref{eq:cosine_schedule}
    \State \textbf{update} $\vz^{(k+1)} = \texttt{AdamStep}(\vz^{(k)};\mathcal{F}_k, \nu_k)$
\EndFor

\State \Return $\vz^{(\texttt{maxit})}$
\end{algorithmic}

\begin{tikzpicture}[overlay, remember picture]
\draw[decorate, decoration={brace, amplitude=5pt, raise=5pt}]
(1.4,7.0) -- (1.4,3.5)
node[midway, xshift=1cm] {$\mathcal{G}(\vz^{(k)})$};
\end{tikzpicture}
\end{algorithm}

\subsection{Network Architecture and Training}

The core component of the DDIM generative framework is a denoising network $f_\Theta(\vx_t, t)$, whose role is to reverse the forward diffusion process by predicting the noise component added to the input $\vx_t$ at diffusion step $t$, as described in Algorithm~\ref{alg:RD-DGP}.

The network $f_\Theta$ is implemented using a UNet architecture. Conditioning on the diffusion timestep $t$ is achieved via a sinusoidal positional embedding~\cite{vaswani2017attention}, which is upsampled to match the spatial resolution of the input and concatenated with $\vx_t$ along the channel dimension. No further timestep conditioning is applied within the network. In our experiments, this simple concatenation strategy proved sufficient to learn an effective generative prior for CT images.

The UNet follows a standard encoder-decoder structure with skip connections. Each resolution level consists of convolutional residual blocks employing Group Normalization and GELU activations to take into account for the limited batch size, which has been set to $16$ due to computational constraints. Spatial downsampling is performed using strided convolutions, while upsampling is implemented via nearest-neighbor interpolation followed by convolution. To balance representational capacity and memory efficiency, self-attention layers are included only at the deepest resolution levels of the network. More precisely, our model comprises four resolution levels with two residual blocks each, and channel widths $(64, 128, 256, 512)$. \\
The model is trained in a self-supervised manner using the standard DDPM/DDIM objective. At each training iteration, a clean image $\vx_0 \sim p_{data}(\vx)$ is corrupted with Gaussian noise at a randomly sampled diffusion timestep $t \sim \mathcal{U}\{1, \dots, T\}$, yielding $\vx_t$ according to the forward diffusion process defined in~\eqref{eq:forward_diffusion}, i.e. $\vx_t = \sqrt{\alpha_t} \vx_0 + \sqrt{1 - \alpha_t} \vepsilon_t$. The network is trained to predict the injected noise $\vepsilon_t \sim \mathcal{N}(\boldsymbol{0}, \boldsymbol{I})$ by minimizing the Denoising Score Matching (DSM) \cite{song2020score} loss:

\begin{align}
\mathcal{L}_{\text{DSM}}(\Theta) = 
\mathbb{E}_{\vx_0 \sim p_{data}}
\mathbb{E}_{t \sim \mathcal{U}[1, T]}
\mathbb{E}_{\vepsilon_t \sim \mathcal{N}(\boldsymbol{0}, \boldsymbol{I})}
\left[
|| f_\Theta(\vx_t, t) - \vepsilon_t ||^2
\right].
\end{align} 
The generative model is trained on axial chest CT slices extracted from the Mayo Clinic Low-Dose CT dataset. Due to the limited size of this dataset, we adopt several strategies to mitigate overfitting and improve generalization. In particular, we rely on the MONAI framework \cite{cardoso2022monai} for both network implementation and data augmentation, leveraging its optimized components for medical imaging applications.
Data augmentation is applied on-the-fly during training and includes random rotations, elastic deformations, affine transformations, and additive Gaussian noise, in addition to standard intensity normalization and resizing. These augmentations are designed to increase variability in anatomical appearance while preserving the underlying CT image characteristics, thereby improving the robustness of the learned generative prior.

To improve training stability, the input data are normalized to the range $[-1,1]$ before being processed by the model, according to
\begin{align}\label{eq:normalization}
\bar{\vx}_0 = 2 \vx_0 - 1.
\end{align}
Since negative pixel values are not physically meaningful in CT imaging, each generated sample is de-normalized after the diffusion process and before application of the operator $\vK$. Specifically, the generated output $\vx_0 = \mathcal{G}(\vz)$ is first clipped to the interval $[-1,1]$ to account for potential numerical drift during generation and then mapped back to the original intensity range by inverting \eqref{eq:normalization}. \\
Training is performed using PyTorch on an NVIDIA A4000 GPU with 16\,GB of VRAM. The model is trained for $1000$ epochs with a batch size of $16$ and a learning rate of $10^{-4}$, using the AdamW optimizer with standard hyperparameters. The forward diffusion process employs a cosine noise schedule with $T=1000$ diffusion steps.

\section{Results and Discussion \label{sec:numexp}}
We evaluate the proposed RD-DGP method across different sparse acquisition geometries. It is important to note that the network is trained only once, and the resulting weights are reused across all experiments.

\begin{figure}[tbhp]
    \centering
    \setlength{\tabcolsep}{1pt}
    \resizebox{\linewidth}{!}{
    \begin{tabular}{cccccc}
    & \textbf{True} & \textbf{DPS} & \textbf{DiffPIR} & \textbf{DDRM} & \makecell{\textbf{RD-DGP}\\\textbf{(Our)}} \\
    \rotatebox{90}{\hspace{5px}\textbf{Angles = $30$}}&
    \imgwithzoom{width=0.2\linewidth}{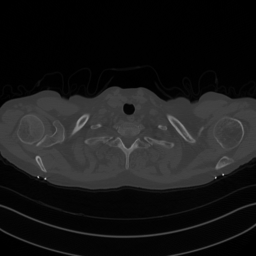}{0.5}{0.4}{0.2}{0.3}{1}&
    \imgwithzoom{width=0.2\linewidth}{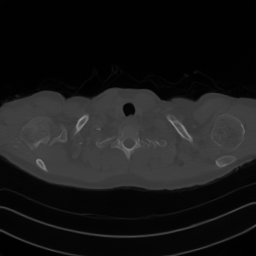}{0.5}{0.4}{0.2}{0.3}{0}&
    \imgwithzoom{width=0.2\linewidth}{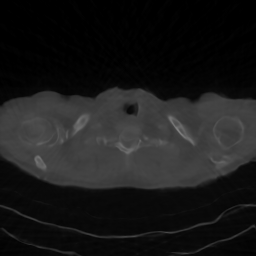}{0.5}{0.4}{0.2}{0.3}{0}&
    \imgwithzoom{width=0.2\linewidth}{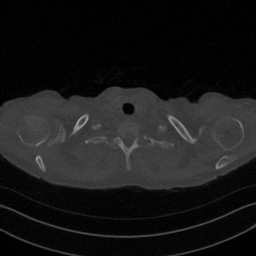}{0.5}{0.4}{0.2}{0.3}{0}&
    \imgwithzoom{width=0.2\linewidth}{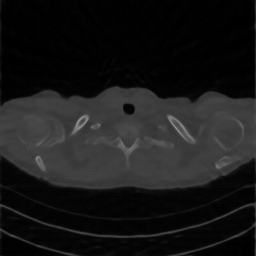}{0.5}{0.4}{0.2}{0.3}{0} \\

    \rotatebox{90}{\hspace{5px}\textbf{Angles = $45$}}&
    \imgwithzoom{width=0.2\linewidth}{imgs/results/C081-35_30_DGP/clean_phantom.png}{0.5}{0.4}{0.2}{0.3}{0}&
    \imgwithzoom{width=0.2\linewidth}{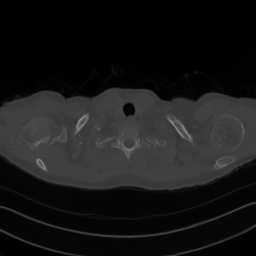}{0.5}{0.4}{0.2}{0.3}{0}&
    \imgwithzoom{width=0.2\linewidth}{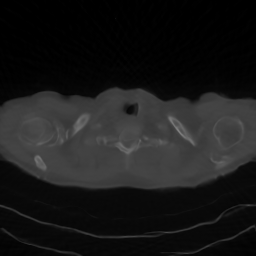}{0.5}{0.4}{0.2}{0.3}{0}&
    \imgwithzoom{width=0.2\linewidth}{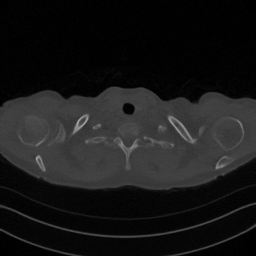}{0.5}{0.4}{0.2}{0.3}{0}&
    \imgwithzoom{width=0.2\linewidth}{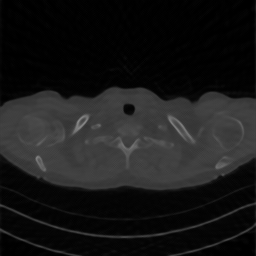}{0.5}{0.4}{0.2}{0.3}{0} \\
    
    \rotatebox{90}{\hspace{5px}\textbf{Angles = $60$}}&
    \imgwithzoom{width=0.2\linewidth}{imgs/results/C081-35_30_DGP/clean_phantom.png}{0.5}{0.4}{0.2}{0.3}{0}&
    \imgwithzoom{width=0.2\linewidth}{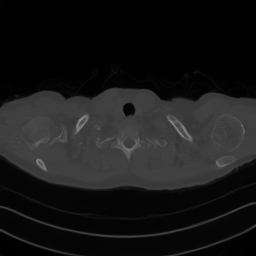}{0.5}{0.4}{0.2}{0.3}{0}&
    \imgwithzoom{width=0.2\linewidth}{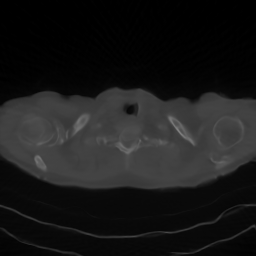}{0.5}{0.4}{0.2}{0.3}{0}&
    \imgwithzoom{width=0.2\linewidth}{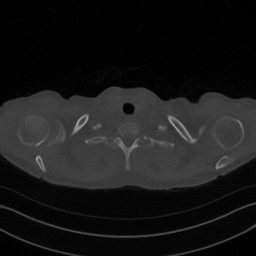}{0.5}{0.4}{0.2}{0.3}{0}&
    \imgwithzoom{width=0.2\linewidth}{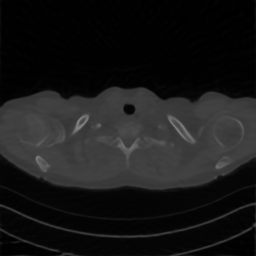}{0.5}{0.4}{0.2}{0.3}{0} \\
    \end{tabular}}
    \caption{Visual comparison of different reconstruction  methods for Sample \texttt{C081-35} across three sparse-view configurations ($n_\alpha = 30, 45, 60$). The red box indicates the zoomed region shown in the inset.}
    \label{fig:qualitative_comparison_35}
\end{figure}

\begin{figure}[tbhp]
    \centering
    \setlength{\tabcolsep}{1pt}
    \resizebox{\linewidth}{!}{
    \begin{tabular}{cccccc}
    & \textbf{True} & \textbf{DPS} & \textbf{DiffPIR} & \textbf{DDRM} & \makecell{\textbf{RD-DGP}\\\textbf{(Our)}} \\
    \rotatebox{90}{\hspace{5px}\textbf{Angles = $30$}}&
    \imgwithzoom{width=0.2\linewidth}{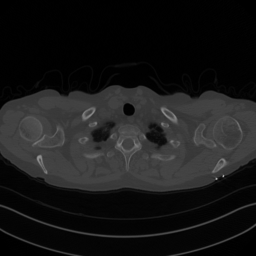}{0.15}{0.45}{0.2}{0.3}{1}&
    \imgwithzoom{width=0.2\linewidth}{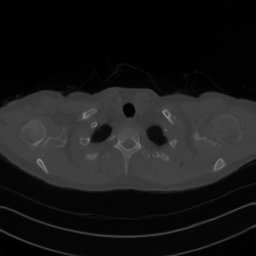}{0.15}{0.45}{0.2}{0.3}{0}&
    \imgwithzoom{width=0.2\linewidth}{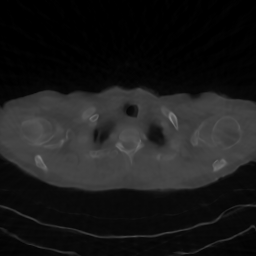}{0.15}{0.45}{0.2}{0.3}{0}&
    \imgwithzoom{width=0.2\linewidth}{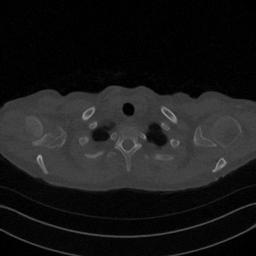}{0.15}{0.45}{0.2}{0.3}{0}&
    \imgwithzoom{width=0.2\linewidth}{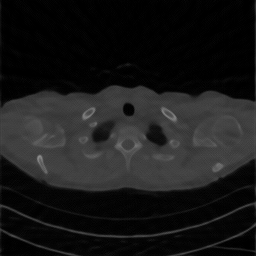}{0.15}{0.45}{0.2}{0.3}{0} \\

    \rotatebox{90}{\hspace{5px}\textbf{Angles = $45$}}&
    \imgwithzoom{width=0.2\linewidth}{imgs/results/C081-45_30_DGP/clean_phantom.png}{0.15}{0.45}{0.2}{0.3}{0}&
    \imgwithzoom{width=0.2\linewidth}{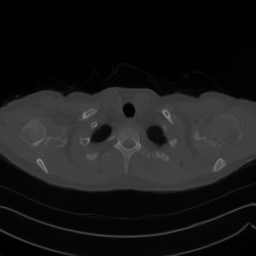}{0.15}{0.45}{0.2}{0.3}{0}&
    \imgwithzoom{width=0.2\linewidth}{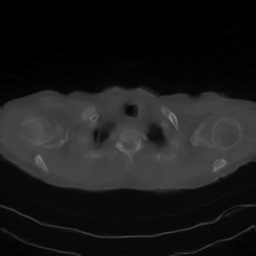}{0.15}{0.45}{0.2}{0.3}{0}&
    \imgwithzoom{width=0.2\linewidth}{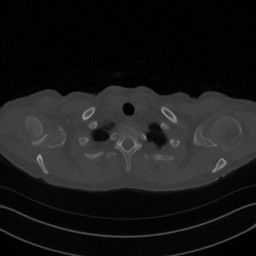}{0.15}{0.45}{0.2}{0.3}{0}&
    \imgwithzoom{width=0.2\linewidth}{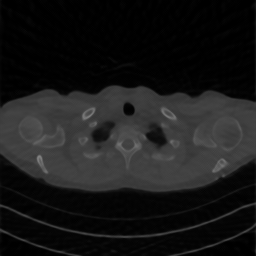}{0.15}{0.45}{0.2}{0.3}{0} \\
    
    \rotatebox{90}{\hspace{5px}\textbf{Angles = $60$}}&
    \imgwithzoom{width=0.2\linewidth}{imgs/results/C081-45_30_DGP/clean_phantom.png}{0.15}{0.45}{0.2}{0.3}{0}&
    \imgwithzoom{width=0.2\linewidth}{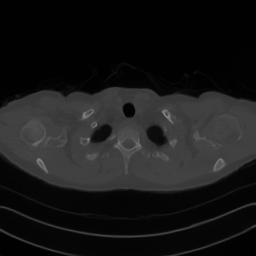}{0.15}{0.45}{0.2}{0.3}{0}&
    \imgwithzoom{width=0.2\linewidth}{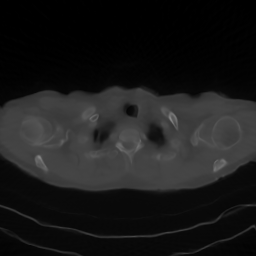}{0.15}{0.45}{0.2}{0.3}{0}&
    \imgwithzoom{width=0.2\linewidth}{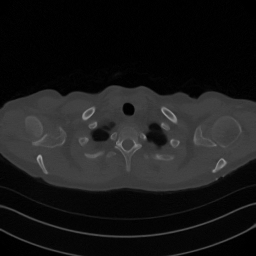}{0.15}{0.45}{0.2}{0.3}{0}&
    \imgwithzoom{width=0.2\linewidth}{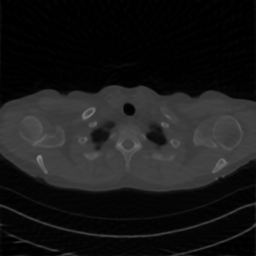}{0.15}{0.45}{0.2}{0.3}{0} \\
    \end{tabular}}
    \caption{Visual comparison of different reconstruction methods for Sample \texttt{C081-45} across three sparse-view configurations ($n_\alpha = 30, 45, 60$). The red box indicates the zoomed region shown in the inset.}
    \label{fig:qualitative_comparison_45}
\end{figure}

\begin{figure}[tbhp]
    \centering
    \setlength{\tabcolsep}{1pt}
    \resizebox{\linewidth}{!}{
    \begin{tabular}{cccccc}
    & \textbf{True} & \textbf{DPS} & \textbf{DiffPIR} & \textbf{DDRM} & \makecell{\textbf{RD-DGP}\\\textbf{(Our)}} \\
    \rotatebox{90}{\hspace{5px}\textbf{Angles = $30$}}&
    \imgwithzoom{width=0.2\linewidth}{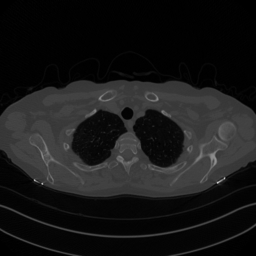}{0.83}{0.45}{0.2}{0.3}{1}&
    \imgwithzoom{width=0.2\linewidth}{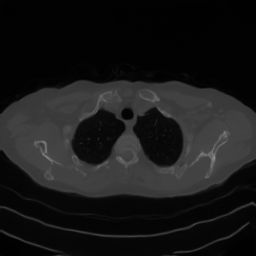}{0.83}{0.45}{0.2}{0.3}{0}&
    \imgwithzoom{width=0.2\linewidth}{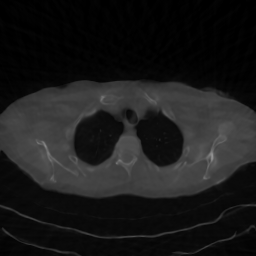}{0.83}{0.45}{0.2}{0.3}{0}&
    \imgwithzoom{width=0.2\linewidth}{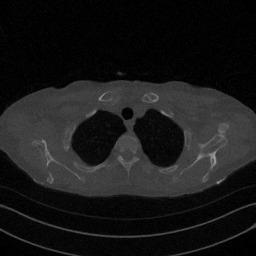}{0.83}{0.45}{0.2}{0.3}{0}&
    \imgwithzoom{width=0.2\linewidth}{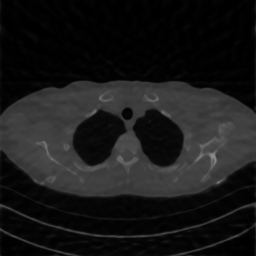}{0.83}{0.45}{0.2}{0.3}{0} \\

    \rotatebox{90}{\hspace{5px}\textbf{Angles = $45$}}&
    \imgwithzoom{width=0.2\linewidth}{imgs/results/C081-79_30_DGP/clean_phantom.png}{0.83}{0.45}{0.2}{0.3}{0}&
    \imgwithzoom{width=0.2\linewidth}{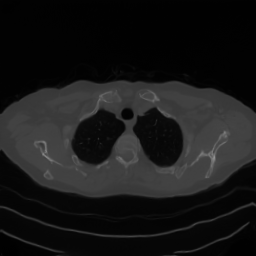}{0.83}{0.45}{0.2}{0.3}{0}&
    \imgwithzoom{width=0.2\linewidth}{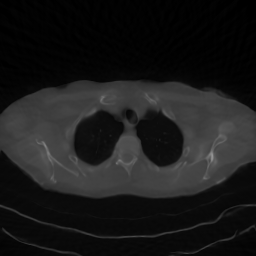}{0.83}{0.45}{0.2}{0.3}{0}&
    \imgwithzoom{width=0.2\linewidth}{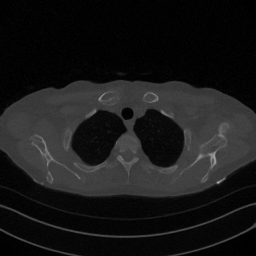}{0.83}{0.45}{0.2}{0.3}{0}&
    \imgwithzoom{width=0.2\linewidth}{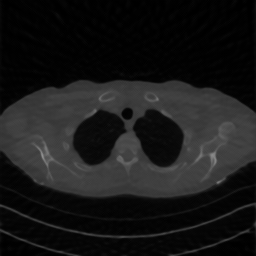}{0.83}{0.45}{0.2}{0.3}{0} \\
    
    \rotatebox{90}{\hspace{5px}\textbf{Angles = $60$}}&
    \imgwithzoom{width=0.2\linewidth}{imgs/results/C081-79_30_DGP/clean_phantom.png}{0.83}{0.45}{0.2}{0.3}{0}&
    \imgwithzoom{width=0.2\linewidth}{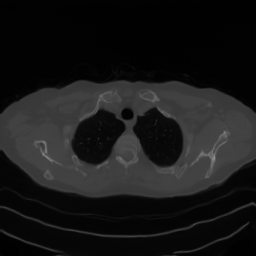}{0.83}{0.45}{0.2}{0.3}{0}&
    \imgwithzoom{width=0.2\linewidth}{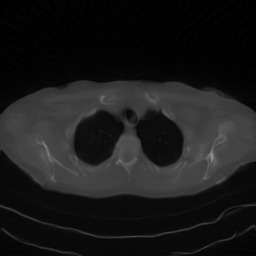}{0.83}{0.45}{0.2}{0.3}{0}&
    \imgwithzoom{width=0.2\linewidth}{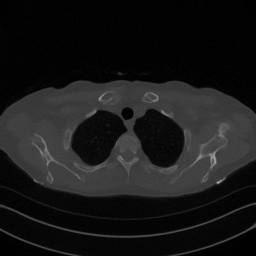}{0.83}{0.45}{0.2}{0.3}{0}&
    \imgwithzoom{width=0.2\linewidth}{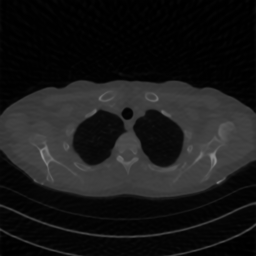}{0.83}{0.45}{0.2}{0.3}{0} \\
    \end{tabular}}
    \caption{Visual comparison of different reconstruction methods for Sample \texttt{C081-79} across three sparse-view configurations ($n_\alpha = 30, 45, 60$). The red box indicates the zoomed region shown in the inset.}
    \label{fig:qualitative_comparison_79}
\end{figure}

We extensively compare the proposed algorithm against existing diffusion-based methods for solving inverse problems: the Diffusion Posterior Sampling (DPS) \cite{chung2022diffusion}, the Diffusion Prior-based Plug-and-Play Image Restoration (DiffPIR) \cite{zhu2023denoising} and the Denoising Diffusion Restoration Model (DDRM) \cite{kawar2022denoising}. Finally, we present a brief ablation study with respect to the modifications introduced relative to the DMPlug algorithm, and we therefore also compare our results with the original proposal in~\cite{wang2024dmplug}.  \\
To compare the generalization performance of the considered methods, we test them on three different test set images, \texttt{C081-35}, \texttt{C081-45}, and \texttt{C081-79}, and we measure the reconstruction quality both via qualitative and quantitative investigation using classical metrics such as the Mean Squared Error (MSE), Peak Signal-to-Noise Ratio (PSNR), and Structural Similarity Index (SSIM). We explore three angular regimes, namely $n_\alpha \in \{30,45,60\}$ angles. \\
All the results are displayed in Table \ref{tab:recon_results_by_angles}.
The DPS and DiffPIR methods exhibit poor performance on all three tests, whereas DDRM and RD-DGP exhibit very similar performance in terms of the considered metrics. In particular, RD-DGP performs slightly better in the geometries with 40 and 60 projection angles, and slightly worse in the case with 30 angles.\\
A qualitative comparison can be carried out by examining 
Figures \ref{fig:qualitative_comparison_35}, \ref{fig:qualitative_comparison_45}, and \ref{fig:qualitative_comparison_79}, which show the reconstructions with the zoomed region identified by the red square. The figures confirm the results reported in the tables, as the reconstructions obtained with DPS and DiffPIR contain noticeable errors. In particular, the DiffPIR reconstruction appears highly blurred and poorly defined.
With respect to the images obtained using the two best-performing methods reported in the table, namely DDRM and RD-DGP, we observe that the resulting images exhibit different characteristics. In particular, the reconstructions of image \texttt{C081-35} (Figure \ref{fig:qualitative_comparison_35}) appear quite similar for both methods, In contrast, for images \texttt{C081-45} and \texttt{C081-79}, inspection of the cropped regions shown in the figures reveals that the DDRM reconstruction exhibits incorrect object shapes, whereas this effect is substantially less pronounced in the RD-DGP reconstruction. Additionally, DDRM images appear more contrasted but also noisier.\\
Finally, we emphasize that the figures demonstrate that, even with a very limited number of projection angles, the images retain good quality, in contrast to reconstructions obtained using non-generative methods.
We also note that an angular step larger than three degrees is far from typical clinical practice and corresponds to extreme test scenarios commonly considered in the literature.

\begin{table}[t]
\centering
\caption{Reconstruction results (PSNR $\uparrow$, SSIM $\uparrow$) for different numbers of acquisition angles. Results are reported for three samples.}
\label{tab:recon_results_by_angles}

\setlength{\tabcolsep}{4pt}
\renewcommand{\arraystretch}{1.05}

\begin{tabular}{c l cc cc cc}
& & \multicolumn{2}{c}{\textbf{\texttt{C081-35}}} 
  & \multicolumn{2}{c}{\textbf{\texttt{C081-45}}} 
  & \multicolumn{2}{c}{\textbf{\texttt{C081-79}}} \\
\cmidrule(lr){3-4}\cmidrule(lr){5-6}\cmidrule(lr){7-8}
& \textbf{Method} 
& \textbf{PSNR} & \textbf{SSIM}
& \textbf{PSNR} & \textbf{SSIM}
& \textbf{PSNR} & \textbf{SSIM} \\
\midrule

\multirow{4}{*}{\rotatebox{90}{\hspace{-10px}$n_\alpha = 30$}}
& \textbf{DPS}     
& $33.32$ & $0.8853$ 
& $32.59$ & $0.8773$
& $34.56$ & $0.8597$ \\
& \textbf{DiffPIR} 
& $31.97$ & $0.8395$ 
& $31.95$ & $0.8323$
& $31.17$ & $0.8251$ \\
& \textbf{DDRM}    
& $35.84$ & $0.8825$ 
& $35.83$ & $0.8870$
& $34.41$ & $0.8507$ \\
\cdashlinelr{2-8}
& \textbf{RD-DGP (Our)}  
& $34.48$ & $0.8802$ 
& $34.38$ & $0.8713$
& $34.46$ & $0.8685$ \\
\midrule\midrule

\multirow{4}{*}{\rotatebox{90}{\hspace{-10px}$n_\alpha = 45$}}
& \textbf{DPS}     
& $33.34$ & $0.8868$ 
& $32.31$ & $0.8759$
& $31.63$ & $0.8618$ \\
& \textbf{DiffPIR} 
& $32.28$ & $0.8517$ 
& $32.25$ & $0.8446$
& $31.47$ & $0.8405$ \\
& \textbf{DDRM}    
& $36.76$ & $0.9163$ 
& $36.66$ & $0.9179$
& $35.78$ & $0.8987$ \\
\cdashlinelr{2-8}
& \textbf{RD-DGP (Our)}  
& $37.27$ & $0.9299$ 
& $37.37$ & $0.9300$
& $35.82$ & $0.9143$ \\
\midrule\midrule

\multirow{4}{*}{\rotatebox{90}{\hspace{-10px}$n_\alpha = 60$}}
& \textbf{DPS}     
& $33.60$ & $0.8902$ 
& $32.92$ & $0.8824$
& $31.61$ & $0.8613$ \\
& \textbf{DiffPIR} 
& $32.33$ & $0.8537$ 
& $32.46$ & $0.8477$
& $31.62$ & $0.8483$ \\
& \textbf{DDRM}    
& $37.17$ & $0.9291$ 
& $37.13$ & $0.9305$
& $36.20$ & $0.9144$ \\
\cdashlinelr{2-8}
& \textbf{RD-DGP (Our)}  
& $37.18$ & $0.9431$ 
& $36.33$ & $0.9394$
& $37.38$ & $0.9355$ \\
\bottomrule
\end{tabular}
\end{table}

\begin{figure}[htbp!]
  \centering

  \setlength{\tabcolsep}{0.6pt}
  \renewcommand{\arraystretch}{0.6}

  \newlength{\schedshift}
  \setlength{\schedshift}{-1.2ex}

  \begin{minipage}{0.32\linewidth}
    \centering
    \footnotesize
    \textbf{$n_\alpha = 30$}\par\vspace{1mm}
    \begin{tabular}{@{}c@{\hspace{3pt}}c@{\hspace{2pt}}c c@{}}
      & & \textbf{FBP Init.} & \textbf{Rand Init.} \\
      & \rotatebox{90}{\textbf{Cosine sched.}} &
        \imgmetric{imgs/results/C081-35_30_DGP/reconstruction.png}{34.48}{0.8802}{0.47\columnwidth} &
        \imgmetric{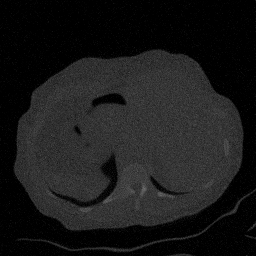}{19.59}{0.4048}{0.47\columnwidth} \\
      & \rotatebox{90}{\hspace{4px}\textbf{No sched.}} &
        \imgmetric{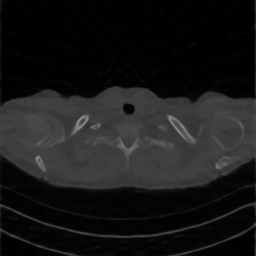}{34.22}{0.8782}{0.47\columnwidth} &
        \imgmetric{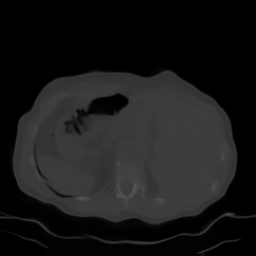}{19.88}{0.4772}{0.47\columnwidth}
    \end{tabular}
  \end{minipage}\hfill
  \begin{minipage}{0.32\linewidth} 
    \centering 
    \footnotesize
    \textbf{$n_\alpha = 45$}\par\vspace{1mm}
    \begin{tabular}{@{}c@{\hspace{3pt}}c@{\hspace{2pt}}c c@{}}
      & & \textbf{FBP Init.} & \textbf{Rand Init.} \\
      & \rotatebox{90}{\textbf{Cosine sched.}} &
        \imgmetric{imgs/results/C081-35_45_DGP/reconstruction.png}{37.27}{0.9299}{0.47\columnwidth} &
        \imgmetric{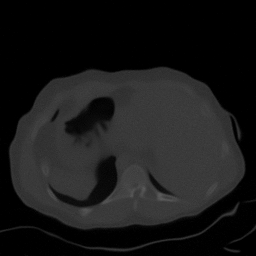}{19.08}{0.4318}{0.47\columnwidth} \\
      & \rotatebox{90}{\hspace{4px}\textbf{No sched.}} &
        \imgmetric{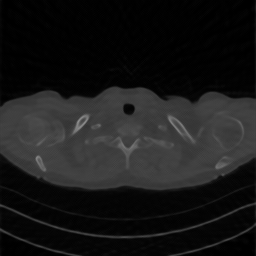}{33.95}{0.8694}{0.47\columnwidth} &
        \imgmetric{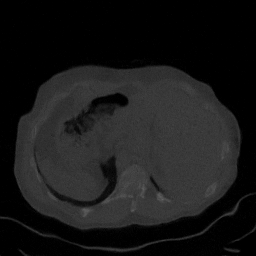}{19.80}{0.4148}{0.47\columnwidth}
    \end{tabular}
  \end{minipage}\hfill
  \begin{minipage}{0.32\linewidth}
    \centering 
    \footnotesize
    \textbf{$n_\alpha = 60$}\par\vspace{1mm}
    \begin{tabular}{@{}c@{\hspace{3pt}}c@{\hspace{2pt}}c c@{}}
      & & \textbf{FBP Init.} & \textbf{Rand Init.} \\
      & \rotatebox{90}{\textbf{Cosine sched.}} &
        \imgmetric{imgs/results/C081-35_60_DGP/reconstruction.png}{37.18}{0.9431}{0.47\columnwidth} &
        \imgmetric{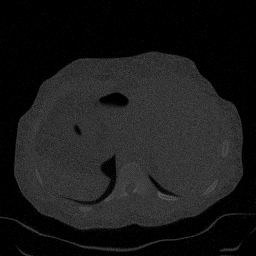}{19.45}{0.4003}{0.47\columnwidth} \\
      & \rotatebox{90}{\hspace{4px}\textbf{No sched.}} &
        \imgmetric{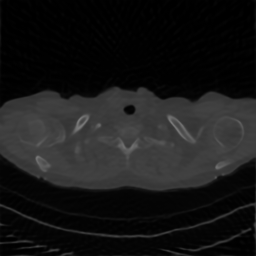}{32.79}{0.8641}{0.47\columnwidth} &
        \imgmetric{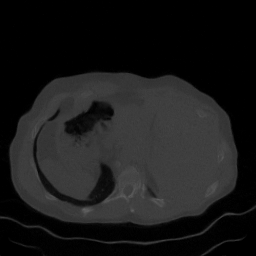}{19.07}{0.4541}{0.47\columnwidth}
    \end{tabular}
  \end{minipage}

  \caption{Ablation study on sample \texttt{C081-35} under sparse-view CT. Each block reports reconstructions (with PSNR/SSIM overlaid) obtained with different combinations of \textbf{FBP initialization} (Yes/No) and \textbf{step-size schedule} (Yes/No). From left to right, the number of projection angles $n_\alpha$ is $30$, $45$, and $60$.}
  \label{fig:ablation}
\end{figure}

Finally, we perform a brief ablation study, the results of which are shown in Figure \ref{fig:ablation}. Here, we consider one test image reconstructed from $n_\alpha=30,45,60$ angles, respectively, and display the results obtained by separately applying the two proposed techniques, namely FBP-based initialization and step-size scheduling in the inversion algorithm. We immediately notice that without the proposed initialization, the reconstructed image obtained by the generative model represents a totally different patient's slice, showing the importance of a correct initialization for such an ill-posed and non-convex optimization problem.
The step-size scheduling, on the other hand, is important for reducing noise and enhancing contrast, both in fine details and in low-contrast regions, but it does not largely affect metrics and is definitely less impactful than the initialization.
Finally, we note that the bottom-right image was obtained using the DMPlug algorithm, which differs from the proposed method in these two modifications, as well as by the inclusion of a regularization function applied to $\G(\vz)$, as in \eqref{eq:RD-DGP}. 
The study confirms the effectiveness of the proposed techniques incorporated in RD-DGP.

\section{Conclusions \label{sec:concl}}

In this work, we address sparse-view CT image reconstruction using recent DDIM-based generative models within a Deep Generative Prior framework, formulated as a regularized least-squares optimization problem. Starting from a recent approach, we enhance its performance through several methodological improvements: initialization of the inverse diffusion process with a coarse filtered backprojection (FBP) reconstruction, the adoption of a reducing step-size schedule strategy in the optimization algorithm, and the incorporation of image-level regularization terms in the objective function. The proposed method is quantitatively and qualitatively compared with state-of-the-art approaches that employ DDIMs for iterative inverse problem solving via data fidelity minimization.

While the results are promising, substantial research efforts are still necessary before generative models can be reliably adopted for solving inverse problems in medical imaging in real-world applications. Significant challenges remain, particularly with respect to computational cost and the robustness and reliability of the reconstructed images. The present work should therefore be regarded as a preliminary step in this direction, contributing to the broader ongoing research aimed at assessing and improving the practical applicability of generative approaches in medical imaging.

\section{Acknowledgements and funding}

This work has been partially supported by the PRIN 2022 project ``STILE: Sustainable Tomographic Imaging with Learning and rEgularization'', project  code: 20225STXSB, funded by the European Commission under the NextGeneration EU programme, and by GNCS - Gruppo Nazionale per il Calcolo Scientifico on the project ``Ottimizzazione bilivello per la gestione automatica del buffer in problemi di Continual Learning``, INdAM GNCS Project, grant code CUP E53C24001950001.

\section{Ethics statements}

This study did not require ethical approval as it exclusively utilized previously published, publicly available dataset repositories of images. No human subjects were directly involved, and no new data collection was performed.

\bibliographystyle{elsarticle-num} 
\bibliography{biblio}

\end{document}